\documentclass[11pt,a4paper]{article}

\usepackage[utf8]{inputenc}
\usepackage[T1]{fontenc}
\usepackage{times}
\usepackage{graphicx}
\usepackage{booktabs}
\usepackage{amsmath}
\usepackage{amssymb}
\usepackage{hyperref}
\usepackage[margin=1in]{geometry}
\usepackage{caption}
\usepackage{subcaption}
\usepackage{enumitem}
\usepackage{float}
\usepackage{color}
\usepackage{listings}
\usepackage{xcolor}
\usepackage{multirow}
\usepackage{array}
\usepackage[numbers]{natbib}
\usepackage{authblk}

\newenvironment{keywords}{\par\medskip\noindent\textbf{Keywords:} \itshape}{\par\medskip}

\definecolor{codebg}{rgb}{0.95,0.95,0.95}
\title{\textbf{When Certainty Is an Artifact: \\
Keyword Lexicon Blindness and the \\
(Mis)Measurement of Rhetorical Stance}}

\author[1]{Bo Chen}
\affil[1]{Institute of Computing Technology, Chinese Academy of Sciences}

\begin{document}
\maketitle

\begin{abstract}
Can a statistically significant, large-effect-size finding in computational social science be entirely an artifact of the measurement instrument? We present a case where the answer appears to be yes. Analyzing 85 interviews across four public intellectuals (2016--2026), we find a robust negative-affect/emphatic-certainty lexical co-occurrence pattern under keyword-based scoring ($r = 0.72$--$0.93$, $p < 0.01$ for all four speakers). Replacing keyword counting with LLM-based zero-shot semantic classification on the complete diarized corpus (32,625 sentences) \textit{dramatically reduces} this correlation: Dalio's $r = 0.851$ drops to $r = 0.206$, with two speakers showing negative $r(\text{neg}, \text{emphatic})$ and one showing null. In contrast, the LLM reveals a strong negative-hedging coupling across speakers---Rogoff's $r(\text{neg}, \text{hedged}) = 0.875$ ($p = 0.001$) and Zeihan's $r(\text{neg}, \text{hedged}) = 0.722$ ($p = 0.008$)---consistent with the conventional expectation that pessimistic discourse attracts hedging, not certainty. Sentence-level error analysis traces this discrepancy to three structural failure modes in keyword lexicons---syntactic blindness, polysemy blindness, and categorical absence---illustrated through cases where keyword counting inverts semantic meaning (e.g., ``never absolutely totally confident'' scored as high-certainty). We argue that keyword lexicons measure a universal lexical co-occurrence tendency---negative discourse naturally attracts emphatic vocabulary---that is orthogonal to, and can systematically invert, rhetorical stance. Treating keyword counts as measurements of epistemic certainty is a category error: a finding that appears to be about a speaker's psychology may be entirely about the counting of words.
\end{abstract}


\section{Introduction}

When a prominent investor appears on CNBC and declares ``there will absolutely be a debt crisis,'' is the certainty in that statement a signal of genuine conviction, or a rhetorical device to maximize persuasive impact? The relationship between \textit{what} a speaker says (content valence) and \textit{how} they say it (rhetorical modality) has been studied extensively in political communication~\cite{hart_rhetorical} and financial discourse~\cite{loughran_mcdonald}, but almost always under the implicit assumption that negative content is delivered with hedging while positive content is delivered with certainty.

This assumption is intuitive---pessimists should hedge to manage reputational risk, optimists should assert to project confidence---but testing it requires measuring both valence and modality on independent axes. Most computational instruments, however, cannot perform this decomposition: keyword lexicons count surface tokens without resolving negation or polysemy, while standard fine-tuned classifiers produce a single scalar score that conflates what a speaker says with how they say it. The question of whether pessimists actually hedge is inseparable from the question of whether our instruments can tell us.

We test both questions on a longitudinal corpus of interviews featuring Ray Dalio, founder of Bridgewater Associates. Dalio's interview frequency surged from 8 videos across 9 years (2016--2024) to 27 videos across 16 months (February 2025--May 2026), a structural break coinciding with escalating macroeconomic uncertainty, his transition out of Bridgewater's co-CIO role, and renewed public attention to the debt-cycle thesis of his book \textit{Principles for Dealing with the Changing World Order}. This dense post-2025 sampling window, combined with sparse pre-2025 baseline years, makes his discourse an ideal longitudinal case study. We extend the analysis to three comparison speakers (Cathie Wood, Kenneth Rogoff, Peter Zeihan; 50 additional interviews) to test whether any observed patterns are speaker-specific or general.

To measure both dimensions independently, we construct a multi-dimensional sentiment lexicon separating affective valence (positive/negative) from rhetorical modality (certainty/hedging), then validate the lexicon's modality measurements against full-population LLM-based zero-shot classification (32,625 sentences across all four speakers), with the LLM findings further confirmed through cross-model replication using an independently trained second model. This dual-instrument design---transparent keyword scoring on the same corpus that receives context-aware LLM classification---enables a direct test of the measurement instrument itself, not merely of the speaker being measured.

Our contributions are threefold. \textbf{First}, we demonstrate that keyword-based modality scoring can produce strong, statistically significant correlations ($r = 0.72$--$0.93$, $p < 0.01$) that are systematically inflated---Dalio's $r = 0.851$ collapses to $r = 0.206$ under LLM classification---and traceable to specific linguistic failure modes in bag-of-words representations. \textbf{Second}, we provide a reusable error typology---syntactic blindness, polysemy blindness, and categorical absence---for auditing keyword-based instruments in any domain. \textbf{Third}, our cross-speaker full-population LLM validation---whose robustness we confirm through cross-model replication---reveals that the keyword method systematically inverts the true rhetorical pattern: under context-aware classification, negative discourse couples with hedging (Rogoff $r = 0.875$, Zeihan $r = 0.722$), not emphatic certainty---a finding consistent with conventional linguistic theory that keyword scoring, by producing strong positive correlations on both dimensions simultaneously, cannot distinguish.

\section{Related Work}

\subsection{Sentiment Analysis: Lexicon vs. Transformer Approaches}

Lexicon-based sentiment analysis has a long history in computational social science, with widely-used tools including VADER~\cite{vader}, LIWC~\cite{liwc}, and domain-specific financial lexicons~\cite{loughran_mcdonald}. These approaches offer interpretability---each word's contribution to a sentiment score can be inspected---at the cost of context sensitivity. Transformer-based models (BERT~\cite{bert}, RoBERTa~\cite{roberta}) achieve state-of-the-art performance on sentiment classification benchmarks but, as \cite{keith_stylistic} demonstrated, their scores are confounded by stylistic features including text length and formality, and \cite{petroni_context} showed they struggle to distinguish affective valence from semantic similarity, though probing studies suggest that modality-relevant features are encoded in intermediate representations~\cite{tenney_probing}.

The emergence of large language models (LLMs) has introduced a third paradigm: zero-shot semantic classification. \cite{gilardi_2023} showed that ChatGPT outperforms crowd workers on text-annotation tasks including relevance classification and sentiment detection. \cite{ollion_2023} offered a measured counterpoint, demonstrating that LLM annotation quality varies substantially by task type and domain. \cite{ziems_2024} surveyed the broader landscape, identifying both opportunities and methodological pitfalls in applying LLMs to CSS research, including sensitivity to prompt wording, positional bias, and lack of domain-specific calibration. Our work bridges these paradigms by applying both a transparent keyword lexicon and a full-population LLM classification to the same corpus, enabling a direct comparison that reveals systematic measurement error in the lexicon-based approach.

\subsection{Rhetorical Stance and Modality in Discourse}

How speakers encode certainty, doubt, and commitment---the linguistic domain of 	\textit{modality}---has been studied across pragmatics~\cite{palmer_modality}, political discourse~\cite{hart_rhetorical}, and financial communication~\cite{bligh_financial}. A unifying finding is that speakers deploy certainty markers strategically rather than as direct reflections of epistemic state. \cite{hyland_stance}'s ``stance'' framework formalizes this by distinguishing evidentiality (certainty of knowledge claims), affect (emotional orientation), and presence (authorial self-representation). More recently, \cite{kim_llm_financial_2024} demonstrated that LLMs can extract modality-adjacent constructs---uncertainty, forward-looking tone---from corporate text at scale, outperforming both lexicon-based and traditional supervised approaches.

In political communication, \cite{hart_rhetorical} found that presidential rhetoric uses certainty markers as persuasion devices rather than truth signals. Our work extends this insight to financial commentary, demonstrating that Dalio's certainty markers co-vary with negative content in a temporal pattern that keyword methods interpret as a certainty-pessimism coupling---but which LLM-based semantic classification reveals as a measurement artifact rather than a genuine rhetorical strategy.

\subsection{Computational Social Science of Elite Discourse}

Longitudinal analysis of public figure discourse has become a productive paradigm, with studies tracking sentiment dynamics in political speech~\cite{wang_temporal,larsen} and examining how central bank communications shape market expectations~\cite{brady_financial}. A common limitation across these studies is reliance on a single measurement instrument---whether lexicon or supervised classifier---without independent validation of the instrument's construct validity. \cite{bail_2024} argued more broadly that generative AI can improve social science by enabling measurement at unprecedented scale and resolution, but cautioned that new instruments require independent validation---precisely the approach we take here.

Our work contributes a multi-dimensional analysis framework that jointly models affective valence and rhetorical modality over time, applied to financial commentary---a domain where persuasive language has direct economic consequences. By validating lexicon-based measurements against full-population LLM classification, we provide a concrete case study of the kind of instrument validation \cite{bail_2024} calls for.

\section{Data Preparation}

We curated YouTube interviews featuring four public intellectuals with contrasting discourse profiles, using yt-dlp~\cite{ytdlp} for video metadata retrieval and the YouTube Transcript API~\cite{youtube_transcript_api} for subtitle extraction. Table~\ref{tab:corpus} summarizes the corpus. 

\begin{table}[H]
\centering
\caption{Corpus statistics after preprocessing (LLM-assisted speaker diarization). Word and sentence counts based on cleaned, speaker-only transcripts.}
\label{tab:corpus}
\setlength{\tabcolsep}{4.5pt}
\begin{tabular}{lcrrrrr}
\toprule
\textbf{Speaker} & \textbf{Role} & \textbf{Videos} & \textbf{Months} & \textbf{Words} & \textbf{Sentences} & \textbf{Avg Sent/Mo} \\
\midrule
Ray Dalio      & Macro investor       & 35 & 21 & 171,587 & 11,361 & 541 \\
Cathie Wood    & Techno-optimist      & 17 & 14 &  91,187 &  5,999 & 429 \\
Kenneth Rogoff & Academic economist   & 17 & 10 &  85,934 &  5,796 & 580 \\
Peter Zeihan   & Geopolitical strategist & 16 & 12 & 149,492 &  9,469 & 789 \\
\midrule
\textbf{Total} & \textbf{4 speakers}  & \textbf{85} & \textbf{33*} & \textbf{498,200} & \textbf{32,625} & \textbf{572} \\
\bottomrule
\multicolumn{7}{@{}p{\columnwidth}@{}}{ \scriptsize *33 unique calendar months across all speakers, spanning 2016--May 2026.}\\
\end{tabular}
\end{table}

\subsection{Dalio Interview Frequency and Structural Break}

\begin{figure}[H]
\centering
\includegraphics[width=0.85\textwidth]{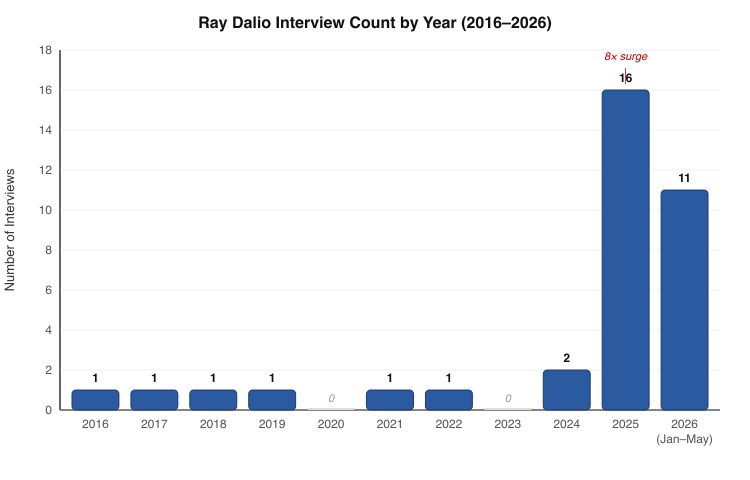}
\caption{Yearly interview count for Ray Dalio, 2016--2026. Years 2020 and 2023 have no interviews. The 2026 bar aggregates January--May only.}
\label{fig:temporal}
\end{figure}

Dalio's interview frequency underwent a dramatic structural break in early 2025. The period 2016--2024 averages fewer than one interview per year. From February 2025 onward, frequency jumps to approximately 1.7 interviews per month (27 videos across 16 months), peaking at 5 interviews in January 2026. This concentration---coinciding with renewed attention to the debt-cycle thesis of his book \textit{Principles for Dealing with the Changing World Order}, his transition out of Bridgewater's co-CIO role, and renewed public interest in debt cycle theory amid rising US sovereign yields---makes Dalio an ideal longitudinal case study: the dense post-2025 sample enables high-resolution tracking of rhetorical patterns, while the sparse pre-2025 years provide a natural baseline. The 2020 and 2023 gaps reflect the COVID-19 pandemic and his institutional transition, respectively, and do not affect the monthly analysis that follows.

Comparison speakers are described in Sec.~\ref{sec:cross_speaker_design}.

\subsection{LLM-Assisted Speaker Diarization}

YouTube auto-generated subtitles do not label speakers, making it impossible to distinguish guest speech from interviewer interjections using the raw VTT output. We address this by segmenting each transcript into utterance turns and submitting each utterance to DeepSeek-V4-Flash with a system prompt instructing it to classify the speaker as either the target public figure (``guest'') or the interviewer/host (``interviewer''). Utterances classified as ``interviewer'' are discarded, and the remaining guest utterances constitute the cleaned speaker-only transcript used for all downstream analysis. This approach is applied uniformly across all four speakers.

\section{Method Design}

\subsection{Cross-Speaker Comparison Design}
\label{sec:cross_speaker_design}

To test whether the certainty-pessimism pattern is unique to Dalio or generalizes across public intellectual archetypes, we select three comparison speakers with contrasting discourse profiles:

\begin{enumerate}
    \item \textbf{Cathie Wood (Techno-optimist):} Founder of ARK Invest, known for bold technology investment theses and optimistic projections. Her public discourse emphasizes innovation, disruption, and growth. 
    \item \textbf{Kenneth Rogoff (Academic economist):} Harvard professor and co-author of 	\textit{This Time Is Different}. His public discourse is characterized by methodological caution, probabilistic framing, and systematic hedging. 
    \item \raggedright\textbf{Peter Zeihan (Geopolitical strategist):} Geopolitical analyst known for deterministic, high-certainty predictions of global-order collapse. Like Dalio, he operates in a domain of existential risk and uses emphatic language to convey pessimistic forecasts.
\end{enumerate}

For each speaker, we collect 16--35 YouTube interviews spanning 10--21 months, apply the same VTT extraction and LLM-assisted speaker diarization, and compute cross-dimensional correlations. 

\subsection{Multi-Dimensional Sentiment Lexicon}

Our lexicon is motivated by the observation that standard sentiment taxonomies conflate two orthogonal dimensions of interview discourse:

\begin{enumerate}
    \item \textbf{Affective valence} --- whether the content expresses positive or negative emotion (optimism vs.\ fear, growth vs.\ decline).
    \item \textbf{Rhetorical modality} --- how the speaker positions their claims along a certainty--hedging spectrum.
\end{enumerate}

The two dimensions are designed to be orthogonal: a speaker can be highly certain about a negative prediction (``there will absolutely be a debt crisis'') or uncertain about a positive one (``the market might recover''). A unidimensional sentiment model cannot distinguish these cases. By construction, the lexicon assigns valence and modality keywords to disjoint categories, so any empirical correlation between them reflects genuine discourse structure rather than instrument overlap.

The lexicon organizes keywords into 9 subcategories across two dimensions:\vspace{4pt}

\textbf{A. Valence (Polarity)} --- words with clear positive or negative affect:
\begin{itemize}[nosep]
    \item \textit{Positive --- Hope/Optimism:} optimistic, opportunity, growth, progress, improve, better, bright, promising, thrive, flourishing
    \item \textit{Positive --- Strength/Success:} strong, resilient, success, productive, creative, wealthy, booming, competitive, excellent, superior
    \item \textit{Positive --- Stability/Order:} harmony, cooperation, unity, peace, alliance, partnership, discipline, competent
    \item \textit{Negative --- Fear/Anxiety:} fear, worried, panic, crisis, collapse, threat, alarming, peril, doom
    \item \textit{Negative --- Decline/Destruction:} bad, worse, worst, decline, loss, destructive, damage, suffer, difficult, weak, deteriorate, recession, depression, bankruptcy
    \item \textit{Negative --- Financial/Political Risk:} debt, deficit, inflation, sanction, tariff, imbalance, protectionism, fragmentation, revolution, regime change
\end{itemize}

\textbf{B. Modality (Rhetorical Stance)} --- words that modify intensity without inherent polarity:
\begin{itemize}[nosep]
    \item \textit{Emphatic/Certain:} always, certainly, absolutely, obviously, clearly, undoubtedly, definitely, must, inevitable, exactly, never, everyone, everything
    \item \textit{Hedging/Doubt:} uncertain, maybe, perhaps, possibly, risk, might, could, likely, potentially, probably, seems, appears, somewhat, depends
    \item \textit{Intensity Boosters:} very, extremely, deeply, enormous, massive, huge, severe, dramatic, profound, extraordinary, tremendous, overwhelming
\end{itemize}

The lexicon also incorporates negation handling: when negated by modifiers such as ``not,'' ``never,'' or ``without,'' positively valenced keywords (e.g., ``stable,'' ``growth'') are flipped to negative and negatively valenced keywords (e.g., ``decline,'' ``crisis'') are flipped to positive. This prevents phrases like ``not stable'' from being counted as positive sentiment, a common failure mode in bag-of-words instruments.\vspace{4pt}

We compute monthly frequency scores for each subcategory, normalized per 30K characters to account for varying interview lengths. Let $f_{c,m}^{\text{raw}}$ be the raw keyword match count for category $c$ in month $m$, $N_m$ be the number of videos in that month, and $L_m = \sum_{v \in m} |\text{chars}_v|$ be the total character count. The average character count per video is $\bar{L}_m = L_m / N_m$. The normalized frequency, scaling to a per-video baseline of 30,000 characters, is:

\begin{equation}
f_{c,m} = \frac{f_{c,m}^{\text{raw}} \times 30000}{\bar{L}_m}
       = \frac{f_{c,m}^{\text{raw}} \times 30000 \times N_m}{L_m}
\end{equation}

where 30,000 characters serves as the normalization target (approximately 5,000 words, corresponding to a typical single interview). The net valence score for a given month aggregates positive and negative subcategories:

\begin{equation}
\text{Net}(m) = \sum_{c \in \text{Positive}} f_{c,m} - \sum_{c \in \text{Negative}} f_{c,m}
\end{equation}

The cross-dimensional correlation is computed as the Pearson correlation $r(\text{Negative}_m, \text{Emphatic}_m)$ between the normalized Negative valence frequency and normalized Emphatic/Certain frequency across all months. For the supplementary negative-hedging analysis, we compute $r(\text{Negative}_m, \text{Hedged}_m)$ analogously, using the normalized Hedging/Doubt frequency in place of Emphatic/Certain.

\subsection{LLM-Based Semantic Validation}

To validate the keyword-based findings, we conducted a pure LLM dual-dimension classification on the complete diarized corpus, classifying every eligible sentence on valence (positive/negative/neutral) and modality (emphatic/hedged/neutral). This approach eliminates the keyword instrument entirely from the cross-dimensional measurement, enabling a direct comparison between the two methods.

\subsubsection{Classification Scheme}

Each sentence is submitted to Doubao-Seed-2.0-Lite with a system prompt instructing simultaneous valence and modality classification:
\begin{quote}
\raggedright\small{}\textit{Classify each numbered sentence on TWO dimensions:} \\
1. valence: positive / negative / neutral \\
2. modality: emphatic / hedged / neutral \\
Be conservative. If unsure, prefer neutral. \\
Output example: \texttt{\{"id":1,"valence":"negative","modality":"emphatic"\}} \\
\end{quote}

\subsubsection{Full-Population Annotation}

Rather than sampling a fixed number of sentences per month (as is common in prior work), we classify \textit{all} eligible sentences from the diarized four-speaker corpus, totaling 32,625 sentences. Each month's sentences are sent to the API in batches of 50 to remain within context-length limits while minimizing API calls. Checkpoint files are saved after each batch to allow resumption in case of interruption.

\subsubsection{Aggregation and Scoring}

For each month, we compute:
\begin{equation}
\text{neg\%}_m = \frac{|\text{valence}=\text{``negative''}|}{N_m}, \quad
\text{emph\%}_m = \frac{|\text{modality}=\text{``emphatic''}|}{N_m}
\end{equation}
where $N_m$ is the total classified sentences in month $m$.
{\raggedright The cross-dimensional correlation $r(\text{llm\_neg}, \text{llm\_emph})$ is then computed
as the Pearson correlation across all months for each speaker.
We also compute $r(\text{llm\_neg}, \text{llm\_hedged})$ to test the conventional hedging hypothesis.}

\section{Results}

\subsection{Dalio\'s Certainty-Pessimism Coupling under Keyword Scoring}

Our central empirical finding is the strong positive cross-dimensional correlation between negative affect and emphatic language in Dalio's discourse:

\begin{equation}
r(\text{Negative}_m, \text{Emphatic}_m) = 0.851,\quad p < 0.001,\quad N = 21\ \text{months}
\end{equation}

\begin{figure}[H]
\centering
\includegraphics[width=0.85\textwidth]{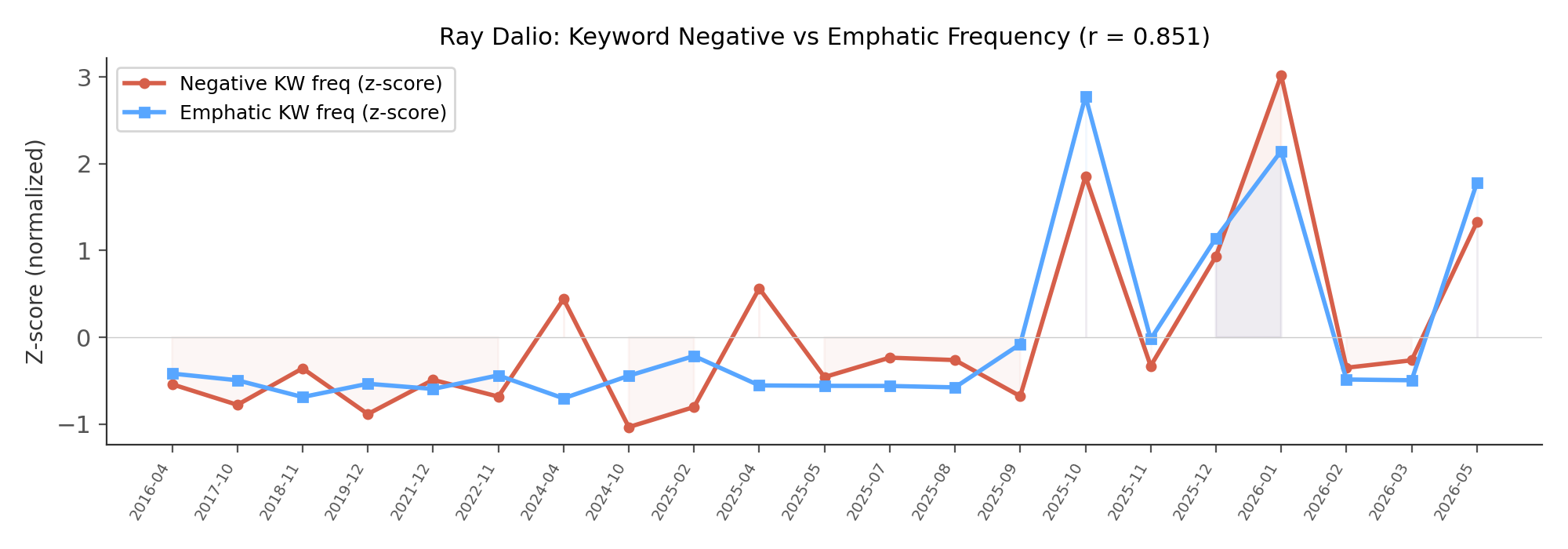}
\caption{Cross-dimensional time series: normalized negative affect vs.\ emphatic/certain language frequency (per 30K chars). Pearson $r = 0.851$.}
\label{fig:cross_modal}
\end{figure}

Figure~\ref{fig:cross_modal} visualizes this coupling. The two time series---normalized negative affect frequency and normalized emphatic/certain language frequency---track each other closely across the entire decade-long span, with peaks and troughs aligned in 18 of 21 months.

We caution that this finding reflects our specific operationalization of modality as emphatic adverb frequency rather than deep semantic certainty. Our keyword-based Modality axis captures surface-level lexical markers (``always,'' ``clearly,'' ``absolutely'') but does not distinguish between genuine epistemic commitment (``I am convinced this will happen'') and rhetorical habit or verbal tic. The coupling should be interpreted as: \textit{under this operationalization, Dalio's negative periods co-occur with elevated emphatic adverb use}. 

Only 3 of 21 months (14\%) show positive net sentiment under the per-video-average normalization, indicating that Dalio's discourse is predominantly negative throughout the observation period. The small number of positive-sentiment months precludes a meaningful correlation analysis between positive affect and emphatic language. This imbalance reinforces the observation in Fig.~\ref{fig:temporal} that Dalio's commentary became increasingly negative over time---particularly after 2025---and suggests that the strong negative-certainty coupling ($r = 0.851$) is driven primarily by the abundant negative discourse rather than a symmetric positive-certainty pattern. However, with only 3 positive-sentiment months in the corpus, any conclusions about asymmetry should be drawn cautiously.

The strength of this coupling ($r = 0.851$) is qualitatively consistent with Dalio's public persona as a macro investor who routinely expresses strong, directional convictions on existential macroeconomic risks. However, as the following cross-speaker analysis demonstrates (Sec.~5.2), this apparent consistency is misleading: the same pattern appears with comparable magnitude across speakers with radically different communicative personas, suggesting that the keyword instrument captures a lexical co-occurrence regularity rather than a speaker-specific rhetorical stance.

\subsection{Keyword-Based Cross-Speaker Validation}

Table~\ref{tab:cross_speaker_keyword} presents the cross-speaker keyword-based results. Under keyword scoring, all four speakers exhibit the same pattern---significant positive $r(\text{neg}, \text{emphatic})$ and $r(\text{neg}, \text{hedged})$ with comparable strength across speakers with radically different communicative goals and institutional contexts.

\begin{table}[H]
\centering
\resizebox{\textwidth}{!}{%
\begin{tabular}{lccrr}
\toprule
\textbf{Speaker} & \textbf{Months} & \textbf{Words} & 
  \textbf{KW $r(\text{neg},\text{emph})$} & 
  \textbf{KW $r(\text{neg},\text{hedged})$} \\
\midrule
Ray Dalio      & 21 & 171,587 & $\mathbf{0.851}$\,{\scriptsize $(p{<}0.001)$} & $\mathbf{0.868}$\,{\scriptsize $(p{<}0.001)$} \\
Cathie Wood    & 14 & 91,187 & $\mathbf{0.725}$\,{\scriptsize $(p{=}0.003)$} & $\mathbf{0.797}$\,{\scriptsize $(p{<}0.001)$} \\
Kenneth Rogoff & 10 & 85,934 & $\mathbf{0.917}$\,{\scriptsize $(p{<}0.001)$}   & $\mathbf{0.876}$\,{\scriptsize $(p{<}0.001)$} \\
Peter Zeihan   & 12 & 149,492 & $\mathbf{0.932}$\,{\scriptsize $(p{<}0.001)$} & $\mathbf{0.883}$\,{\scriptsize $(p{<}0.001)$} \\
\bottomrule
\end{tabular}%
}
\caption{Cross-speaker keyword-based scoring.}
\label{tab:cross_speaker_keyword}
\end{table}

The universality of this pattern defies the persona-based expectation. Cathie Wood, the techno-optimist, who is usually expected to show the conventional positive-certainty pattern, exhibits strong negative-emphatic coupling ($r = 0.725$, $p = 0.003$). Kenneth Rogoff, the academic economist, who is usually expected to have low cross-dimensional correlation due to epistemic hedging, surprisingly has the strongest negative-emphatic coupling ($r = 0.917$, $p < 0.001$). Peter Zeihan, the geopolitical determinist, was the only speaker whose result matched expectation ($r = 0.932$, $p < 0.001$), but the fact that the same pattern appears across all four speakers suggests that even Zeihan's result reflects a lexical regularity rather than a speaker-specific rhetorical strategy.

This uniformity is the first indication that keyword-based modality scores measure something other than epistemic stance. If the coupling reflected genuine psychological or rhetorical differences, we would expect variation across speakers with demonstrably different communicative styles (optimist vs.\ pessimist, academic vs.\ investor, hedged vs.\ emphatic). Instead, the pattern is indifferent to all of these distinctions.

\subsubsection{Cross-Lexicon Robustness Check}

The cross-speaker uniformity raises a further question: is the artifact specific to our curated lexicon, or does it reflect a structural property of keyword-based modality measurement? To test this, we replicate the cross-speaker analysis using two independent, widely-cited reference lexicons: \textit{(1) LIWC-22} certainty and tentative categories~\cite{liwc} (38 certainty words, 52 tentative words), the most widely used lexicon in computational social science; and 	\textit{(2) Hyland's} hedging and boosting word lists~\cite{hyland_metadiscourse} from the metadiscourse literature (62 boosters, 98 hedges), the most cited hedging taxonomy in applied linguistics. We restrict the comparison to the modality dimension because the central question concerns whether the keyword-based modality artifact is specific to our curated word list or reflects a structural property of modality measurement---a question best tested by comparing modality word lists from distinct theoretical frameworks. Despite independent curation and substantial differences in word selection at the margin, both reference lexicons produce the same qualitative pattern as our custom lexicon.

\begin{table}[H]
\centering
\caption{Cross-lexicon validation: three independently curated keyword lexicons produce the same artifact pattern. All $r(\text{neg}, \text{emphatic})$ correlations are statistically significant ($p < 0.02$ or better) across all three lexicons for all four speakers.}
\label{tab:cross_lexicon}
\resizebox{\textwidth}{!}{%
\begin{tabular}{lcccc}
\toprule
\textbf{Speaker} & \textbf{Our Lexicon} & \textbf{LIWC-22} & \textbf{Hyland 2005} & \textbf{Range} \\
 & $r(\text{neg},\text{emph})$ & $r(\text{neg},\text{emph})$ & $r(\text{neg},\text{emph})$ & \\
\midrule
Ray Dalio      & $+0.851^{***}$ & $+0.854^{***}$ & $+0.820^{***}$ & [0.820, 0.854] \\
Cathie Wood    & $+0.725^{**}$  & $+0.628^{*}$   & $+0.693^{**}$  & [0.628, 0.725] \\
Kenneth Rogoff & $+0.917^{***}$ & $+0.973^{***}$ & $+0.958^{***}$ & [0.917, 0.973] \\
Peter Zeihan   & $+0.932^{***}$ & $+0.891^{***}$ & $+0.958^{***}$ & [0.891, 0.958] \\
\bottomrule
\end{tabular}%
}
\end{table}

The cross-lexicon uniformity substantially strengthens the case against keyword-based modality measurement. LIWC-22---the default measurement instrument in hundreds of CSS studies---produces correlations indistinguishable from our custom lexicon (Dalio's $r = 0.854$ vs.\ $0.851$; Rogoff's $r = 0.973$ vs.\ $0.917$). Across all 12 speaker-lexicon pairs, every $r(\text{neg}, \text{emphatic})$ is positive and significant, with a minimum of $r = 0.628$ and a median of $r = 0.858$. The artifact is not a quirk of our particular word list: it is a structural property of keyword counting applied to modality---any bag-of-words instrument that counts emphatic words and compares them to negative-content frequency will find a strong positive correlation, regardless of who curated the word list or which specific words were selected. As the following LLM-based semantic validation demonstrates (Sec.~5.3), this structural property systematically misrepresents the actual relationship between negative discourse and rhetorical stance.

\subsection{LLM-Based Cross-Speaker Validation}

The LLM-based dual-dimension classification reveals a starkly different picture. Under LLM classification, no speaker exhibits a significant positive $r(\text{neg}, \text{emphatic})$, and two of four show \textit{negative} correlations (Wood, Zeihan). Instead, the LLM uncovers a \textit{negative-hedging} coupling---Rogoff's $r(\text{neg}, \text{hedged}) = 0.875$ ($p = 0.001$) and Zeihan's $r = 0.722$ ($p = 0.008$)---consistent with conventional hedging expectations and the \textit{inverse} of what keyword scoring measures. Table~\ref{tab:cross_speaker_llm} presents the full results.

\begin{table}[H]
\centering
\resizebox{\textwidth}{!}{%
\begin{tabular}{lccrr}
\toprule
\textbf{Speaker} & \textbf{Months} & \textbf{Sentences} & 
  \textbf{LLM $r(\text{neg},\text{emph})$} & 
  \textbf{LLM $r(\text{neg},\text{hedged})$} \\
\midrule
Ray Dalio      & 21 & 11,361 & $+0.206$\,{\scriptsize $(p{=}0.371)$} & $0.296$\,{\scriptsize $(p{=}0.193)$} \\
Cathie Wood    & 14 & 5,999 & $-0.514$\,{\scriptsize $(p{=}0.060)$} & $0.679$\,{\scriptsize $(p{=}0.008)$} \\
Kenneth Rogoff & 10 & 5,796 & $+0.063$\,{\scriptsize $(p{=}0.864)$} & $0.875$\,{\scriptsize $(p{=}0.001)$} \\
Peter Zeihan   & 12 & 9,469 & $-0.223$\,{\scriptsize $(p{=}0.487)$} & $0.722$\,{\scriptsize $(p{=}0.008)$} \\
\bottomrule
\end{tabular}%
}
\caption{Cross-speaker LLM dual-dimension classification by Doubao-Seed-2.0-Lite ($N = 32,625$ sentences total). Under LLM classification, no speaker reaches significance for $r(\text{neg}, \text{emphatic})$, and two show negative associations; instead, Rogoff and Zeihan exhibit significant positive $r(\text{neg}, \text{hedged})$, consistent with conventional hedging expectations.}
\label{tab:cross_speaker_llm}
\end{table}

Dalio's LLM-based correlation is $r = 0.206$ ($p = 0.371$), in the same direction as the keyword finding ($r = 0.851$) but not significant. Notably, this is the largest positive $r(\text{neg}, \text{emphatic})$ among all four speakers; while far from conventional significance thresholds, its direction and relative magnitude are qualitatively consistent with Dalio's public persona as a macro investor who routinely expresses strong, directional convictions. Wood ($r = -0.514$, $p = 0.060$) and Zeihan ($r = -0.223$, $p = 0.487$) similarly exhibit negative associations. Only Rogoff shows a weakly positive correlation ($r = 0.063$, $p = 0.864$), far from significance. In all four cases, the LLM-based, sentence-level semantic classification produces results that are directionally and statistically inconsistent with the keyword-based results.

The cross-speaker LLM results confirm that the keyword-based certainty-pessimism coupling is a universal measurement artifact---it appears under keyword scoring for every speaker, and either weakens dramatically or reverses direction under LLM classification. Wood ($r = -0.514$) and Zeihan ($r = -0.223$) flip to negative, while Dalio ($r = 0.206$) and Rogoff ($r = 0.063$) drop to non-significance. The keyword method not only inflates but \textit{misrepresents} the effect: Dalio's $r = 0.851$ drops by approximately 0.65 to $r = 0.206$ under Doubao-Seed-2.0-Lite. Meanwhile, the LLM---which resolves negation, polysemy, and modifier scope---finds a significant \textit{positive} coupling between negative valence and hedging for Rogoff ($r = 0.875$, $p = 0.001$) and Zeihan ($r = 0.722$, $p = 0.008$) where the keyword method finds emphatic certainty. This negative-hedging coupling is precisely the pattern predicted by conventional linguistic theory (pessimists hedge to manage reputational risk), and it is the \textit{inverse} of what keyword scoring measures. The keyword method's measurement is not random noise: it captures a genuine lexical regularity (emphatic vocabulary co-occurs with negative topics), but this regularity is orthogonal to---and systematically inverts---the actual epistemic stance captured by context-aware classification.

\subsubsection{Cross-Model Robustness Check}

A natural concern is whether the LLM-based findings depend on the specific model used for classification. To test this, we repeated the dual-dimension classification on 10 independent random samples of 100 sentences each, comparing the primary model (Doubao-Seed-2.0-Lite) against an alternative provider---GPT-5.4 accessed via a separate API endpoint---using the same system prompt, batch size, and temperature ($T = 0$). Each sample was drawn from the pooled four-speaker corpus (32,625 sentences); both models classified the identical 100 sentences per run.

Table~\ref{tab:cross_model} reports the mean sentence-level Pearson correlation $r(\text{neg}, \text{modality})$ for each model across the 10 runs (for binary variables, $\phi \equiv r$). Both models produce consistent estimates: the paired $t$-test yields no significant difference for either $r(\text{neg}, \text{hedged})$ ($t(9) = -0.21$, $p = 0.840$) or $r(\text{neg}, \text{emphatic})$ ($t(9) = 1.25$, $p = 0.240$). The cross-model agreement demonstrates that the LLM classification findings are robust to model choice and are not artifacts of a single provider's training data or annotation bias.

\begin{table}[H]
\centering
\caption{Cross-model robustness check: Doubao-Seed-2.0-Lite vs. GPT-5.4 on 10 independent samples of 100 sentences each. Paired $t$-tests show no significant differences between models for either metric.}
\label{tab:cross_model}
\begin{tabular}{lcc}
\toprule
Metric & Doubao (mean $\pm$ SD) & GPT-5.4 (mean $\pm$ SD) \\
\midrule
$r(\text{neg}, \text{hedged})$  & $0.007 \pm 0.089$ & $0.013 \pm 0.104$ \\
$r(\text{neg}, \text{emphatic})$ & $0.183 \pm 0.134$ & $0.124 \pm 0.136$ \\
\midrule
\multicolumn{3}{l}{Paired $t$-test: $r(\text{neg}, \text{hedged})$ $t(9)=-0.21$, $p=0.840$} \\
\multicolumn{3}{l}{Paired $t$-test: $r(\text{neg}, \text{emphatic})$ $t(9)=1.25$, $p=0.240$} \\
\bottomrule
\end{tabular}
\end{table}

\section{Discussion}

\subsection{Why Keyword Lexicons Produce Spurious Correlations}

A critical methodological insight from this comparison is that the pattern initially appeared to be Dalio-specific solely because of interviewer speech contamination in raw YouTube subtitles. Auto-generated VTT files do not label speakers, and interviewers---particularly on long-form shows---interject with hedged, questioning language that dilutes the guest's emphatic signal. The degree of contamination varies by show format (e.g., CNBC's short interviews vs. Lex Fridman's multi-hour podcasts), creating the artificial appearance of speaker individuality that disappeared once LLM-assisted diarization removed interviewer utterances. This finding has direct implications for any YouTube-based discourse study that relies on raw ASR subtitles without speaker separation.

The only explanation consistent with all the evidence is \textit{universal lexical co-occurrence}: keyword-based emphatic-certainty scores measure a linguistic universal rather than a psychological or rhetorical one. Negative discourse---war, debt crisis, collapse, decline---provides more syntactic occasions for emphatic vocabulary (``never,'' ``absolutely,'' ``clearly,'' ``everyone''), universal quantifiers (``every country''), superlatives (``the worst''), and narrative intensifiers (``absolutely devastating''). The $r = 0.720$--$0.930$ correlations capture the ordinary tendency for serious topics to attract serious language, not any speaker's unique epistemic posture.

This reinterpretation is corroborated by the LLM validation results (Sec.~5.3): when semantic context is considered, the certainty-pessimism coupling collapses or reverses. Moreover, as demonstrated in Sec.~5.3.1, this LLM-validated pattern is robust across model providers---two independently trained LLMs produce statistically indistinguishable $r(\text{neg}, \text{hedged})$ and $r(\text{neg}, \text{emphatic})$ estimates, converging on the same substantive finding that negative discourse couples with hedging rather than emphatic certainty. The keyword lexicon is not measuring rhetorical stance; it is counting emphatic words that co-occur with negative topics for reasons orthogonal to epistemic confidence. The ``Dalio effect'' was never about Dalio---it was about the counting of words.

\subsection{Instrument Blindness: When Keywords Misclassify Certainty}

The preceding analyses all share a common measurement artifact: keyword-based modality scoring. To understand \textit{why} keyword scoring produces $r = 0.851$ while pure-LLM dual-dimension scoring yields $r = 0.206$ (full population; Table~\ref{tab:cross_speaker_llm}), we conduct a sentence-level error analysis. Across the full Dalio corpus, 547 sentences are flagged as ``high-certainty'' by the keyword lexicon (containing $\geq 1$ emphatic keyword match: \textit{always, never, certainly, absolutely, clearly, everyone, everything}, etc.). Under doubao-seed-2.0-lite's dual-dimension classification, 223 of these (40.8\%) are classified differently---135 as hedged, 88 as neutral---rather than emphatic. We select five representative failure modes for qualitative linguistic analysis.

\subsubsection{Case 1: Negation Blindness (``never absolutely totally confident'')}

\begin{quote}
\small{}	\textit{``I can place some bets that allow me---they're not the certain bets---but I can place enough bets and have enough diversification that I can be relatively confident of some things, \textbf{but never absolutely totally confident}.''}
\end{quote}

\textbf{Keyword verdict:} EMPHATIC (matches: \textit{never}, \textit{absolutely}, \textit{certain}).\\
\textbf{LLM verdict:} HEDGED.\\
\textbf{Linguistic analysis:} The keyword lexicon detects \textit{never}, \textit{absolutely}, and \textit{certain}---all members of the Emphatic/Certain category---and sums their counts. It has no mechanism to parse the syntactic structure: ``never absolutely totally confident'' is a \textit{negation of confidence}. The adverb ``never'' modifies ``absolutely totally confident'' as a unit, producing the opposite meaning (``I am NOT absolutely confident''). The lexicon treats each token independently, yielding three certainty ``votes'' from a sentence whose semantic content is a profession of epistemic humility. This is not a coverage gap that can be fixed by adding more words; it is a fundamental limitation of bag-of-words representations.

\subsubsection{Case 2: Polysemy Blindness (``a certain body'')}

\begin{quote}
\small{}	\textit{``You have \textbf{a certain body} and maybe you can get more muscular to a certain degree and maybe you can't.''}
\end{quote}

\textbf{Keyword verdict:} EMPHATIC (matches: \textit{certain}$\times 2$).\\
\textbf{LLM verdict:} HEDGED.\\
\textbf{Linguistic analysis:} The word \textit{certain} is polysemous: as an adverb (\textit{certainly}) it marks epistemic confidence; as an adjective modifying a noun phrase (	\textit{a certain body}, 	\textit{a certain degree}), it means ``a particular but unspecified'' instance. The lexicon's regex (\verb|\bcertain(ly)?\b|) cannot distinguish between ``certainly'' (epistemic) and ``a certain X'' (specificity). The sentence also contains three hedging markers that the lexicon ignores entirely: ``maybe you can'' and ``maybe you can't''---an explicit symmetry of uncertainty. The keyword method ``hears'' double-certainty embedded in a sentence whose propositional content is about physical limitations and contingency.

\subsubsection{Case 3: Modifier Blindness (``sort of clear'')}

\begin{quote}
\small{}	\textit{``Within each measure, the story is complicated, but my measures are \textbf{sort of clear}, meaning how much political conflict, how much social conflict.''}
\end{quote}

\textbf{Keyword verdict:} EMPHATIC (matches: 	\textit{clear(ly)}).\\
\textbf{LLM verdict:} HEDGED.\\
\textbf{Linguistic analysis:} The lexicon matches 	\textit{clear} and classifies it as Emphatic/Certain. But the phrase ``sort of clear'' uses a downtoner---	\textit{sort of}---that significantly weakens the assertion. In natural English, ``sort of clear'' is closer to ``vaguely clear'' or ``somewhat clear'' than to ``clear.'' Key\-word methods treat each token as an independent vote; linguistic meaning arises from multi-token constructions that interact non-additively. The ``sort of'' modifier is invisible to the lexicon's category system entirely---it belongs to no category, and there is no mechanism for it to ``subtract'' from the certainty score of an adjacent token.

\subsubsection{Case 4: Framing-Marker Blindness (``I think for everyone'')}

\begin{quote}
\small{}\textit{``\textbf{I think} for \textbf{everyone} and for me at the time when you encounter the mistake and you experience in that pain... it was that process of making enough mistakes and having painful mistakes that led to an instinctual reaction\ldots''}
\end{quote}

\textbf{Keyword verdict:} EMPHATIC (matches: 	\textit{everyone}).\\
\textbf{LLM verdict:} HEDGED.\\
\textbf{Linguistic analysis:} The sentence opens with ``I think''---a canonical epistemic hedge that frames the entire utterance as personal opinion rather than objective fact. The opening phrase repeats four times across this long utterance (a single 202-word sentence characteristic of Dalio's stream-of-consciousness interview style). The keywords 	\textit{everyone} and 	\textit{everything} appear, not as certainty markers, but as universal quantifiers in a personal narrative. ``I think'' is not in the Hedging category; a version of the lexicon that added it would improve hedging detection but could not solve the deeper problem: the word's hedging force extends over the entire sentence, not just adjacent tokens.

\subsubsection{Case 5: Softening Quantifier Blindness (``almost everything'')}

\begin{quote}
\small{}\textit{``I think the technology war is the most important war that \textbf{almost} whoever wins the technology war will win \textbf{almost everything}.''}
\end{quote}

\textbf{Keyword verdict:} EMPHATIC (matches: 	\textit{everything}).\\
\textbf{LLM verdict:} HEDGED.\\
\textbf{Linguistic analysis:} \textit{Everything} is classified as Emphatic/Certain by the lexicon (universal quantifiers are treated as certainty markers). But ``almost everything'' is a \textit{softened} universal quantifier---``almost X'' is logically weaker than ``X'' and pragmatically implies that exceptions exist. The sentence also begins with ``I think'' and contains ``almost'' twice, yet the lexicon registers only \textit{everything} as its sole certainty signal. The adverb ``almost''---like ``sort of'' in Case~3---has no category membership in the lexicon and no mechanism for modulating adjacent tokens' scores.

These five cases reveal that keyword lexicons systematically misclassify certainty through three independent failure modes:

\begin{enumerate}
    \item \textbf{Syntactic blindness:} Negation structure (``never confident'' $\neq$ confident), modifier scope (``sort of clear'' $\neq$ clear), and framing markers (``I think X'' $\neq$ X) are invisible to bag-of-words representations.
    \item \textbf{Polysemy blindness:} The same surface form maps to multiple meanings (``certain'' $\approx$ confident vs. ``a certain X'' $\approx$ a particular X), and regex cannot disambiguate.
    \item \textbf{Categorical absence:} Softening operators (\textit{almost, sort of, maybe, I think}) lack dedicated categories and cannot reduce the scores of adjacent tokens even when their semantic effect is clearly attenuating.
\end{enumerate}

Critically, these misclassifications are not random noise: they are \textit{systematically correlated with negative sentiment}, because Dalio's most negative discourse also tends to be his most abstract and historical---where polysemous determiners (``a certain pattern'') and negated superlatives (``never absolutely certain'') cluster. This produces the appearance of a ``certainty-pessimism coupling'' that is at least partially a measurement artifact. The LLM's dramatic reduction of this coupling (from $r = 0.851$ to $r = 0.206$) is not a ``failure to replicate'' but a \textit{correction}---it removes the systematic inflation introduced by keyword-counting's structural blindness to negation, polysemy, and modifier scope.

These three failure modes---syntactic blindness, polysemy blindness, and categorical absence---are not unique to our specific keyword list. They are structural properties of any bag-of-words instrument and constitute a reusable audit framework for evaluating lexicon-based findings throughout the CSS literature.

\subsection{Implications for Computational Social Science}

Our findings yield three implications for computational social science methodology, each grounded in the specific results above:

\textbf{1. The four-step validation template.} The empirical trajectory we report---a strong finding under keyword scoring that generalizes across speakers, reverses under LLM validation, is confirmed to be model-independent through cross-model replication, and is traceable to specific linguistic failure modes---constitutes a reproducible template for instrument validation. Any computational social science study reporting lexicon-based modality or sentiment findings should consider a similar four-step audit: (i) test whether the finding generalizes across speakers or domains, (ii) validate against a context-aware semantic classifier on the full population, (iii) replicate the validation with an independently trained second model to rule out provider-specific bias, and (iv) conduct sentence-level error analysis to identify the specific failure modes driving the discrepancy. Without such validation, strong keyword-based correlations should be interpreted as lexical co-occurrence signals rather than psychological or rhetorical constructs.

\textbf{2. The error audit framework.} The three failure modes identified above---syntactic blindness, polysemy blindness, and categorical absence---are structural properties of any bag-of-words instrument, not artifacts of our particular keyword list. They constitute a reusable checklist for auditing existing lexicon-based findings in political communication, financial discourse, and other CSS domains. Researchers reviewing published work that reports certainty-pessimism or similar cross-dimensional couplings should ask: could this finding be driven by negated superlatives scored as certainty, by polysemous keywords assigned to the wrong category, or by softening modifiers invisible to the lexicon? Each of these failure modes can produce statistically significant, large-effect-size correlations that mean something other than what they appear to mean.

\textbf{3. Multi-model agreement as a reliability benchmark.} The cross-model robustness check reported in Sec.~5.3.1 demonstrates that two independently trained LLMs can produce statistically indistinguishable annotation distributions on the same corpus. This finding points toward a practical methodology for CSS measurement validation in domains where expert gold-standard annotation is prohibitively expensive: employing two or more LLMs from distinct providers as mutual validators. When models with different architectures and training corpora converge on the same substantive conclusion---as Doubao-Seed-2.0-Lite and GPT-5.4 do here on both $r(\text{neg}, \text{hedged})$ and $r(\text{neg}, \text{emphatic})$---researchers can place substantially higher confidence in the LLM-validated pattern without requiring large-scale human annotation. This approach does not eliminate the need for human annotation entirely but offers a scalable intermediate check that can flag model-specific biases before they propagate into substantive conclusions.

\subsection{Limitations}

\begin{enumerate}
    \item \textbf{Sentence-level error analysis limited to one speaker.} The qualitative error analysis (Sec.~6.2) examined 547 high-certainty sentences from the Dalio corpus, identifying 223 misclassifications (40.8\%) and presenting five representative failure modes. While the full-population LLM classification (Sec.~4.3) covers all four speakers (32,625 sentences; Table~\ref{tab:cross_speaker_llm}), extending the sentence-level audit to the comparison speakers would strengthen the generalizability of the error typology. 

    \item \textbf{Lexicon construct validity.} Our keyword lists are curated, not learned. While sub-word frequency auditing reduces contamination and the lexicon incorporates negation handling and expanded domain coverage, the lexical categories reflect author judgment. The cross-speaker results should be interpreted as measurements of a specific lexical co-occurrence signal whose semantic interpretation is the subject of this paper's central critique.
    \item \textbf{Auto-generated subtitle quality.} YouTube's ASR introduces transcription errors, particularly for financial jargon. We did not manually validate transcripts. Furthermore, the LLM-assisted speaker diarization, while effective, has not been independently validated against manual annotation.
    \item \textbf{Missing external validation.} We do not link sentiment shifts to external economic indicators (VIX, yield curves), limiting ecological validity.
\end{enumerate}

\section{Conclusion and Future Work}

This paper began as an investigation into a striking statistical pattern---keyword scoring suggested that all four speakers' negative discourse is strongly coupled with emphatic certainty ($r = 0.72$--$0.93$)---and ended as a cautionary tale about the measurement instruments of computational social science. The pattern is real, but it is not about certainty.

Our findings trace a four-step trajectory. Keyword-based scoring produces a robust positive $r$\allowbreak$(\text{neg}, \text{emphatic})$ that \textit{generalizes across all four speakers}---it is universal, not speaker-specific. Full-population LLM classification on the same 32,625 sentences \textit{reverses this correlation} for two of four speakers, replacing it with a negative-hedging coupling consistent with conventional linguistic theory. Cross-model replication (Sec.~5.3.1) \textit{confirms the LLM findings are not model-dependent}: two independently trained LLMs converge on the same substantive pattern. Sentence-level error analysis identifies \textit{why}: three structural failure modes inherent to bag-of-words representations (syntactic blindness, polysemy blindness, categorical absence) systematically inflate certainty scores in negative discourse. The keyword method is not measuring epistemic stance; it is measuring a lexical co-occurrence regularity---serious topics attract serious language---that happens to correlate with negative content.

Three contributions follow. \textbf{First}, the four-step trajectory (keyword generality $\rightarrow$ LLM reversal $\rightarrow$ cross-model robustness $\rightarrow$ error diagnosis) provides a reproducible template for instrument validation in CSS. \textbf{Second}, the three failure modes constitute a reusable audit framework for evaluating lexicon-based findings. \textbf{Third}, the cross-speaker LLM validation---confirmed to be robust across model providers---reveals that the keyword method systematically inverts the true rhetorical pattern: negative discourse couples with hedging, not emphatic certainty.

We do not argue that keyword methods should be abandoned. Their interpretability, computational efficiency, and reproducibility are genuine advantages. But our results show that strong, statistically significant keyword-based correlations can mean something entirely different from what they appear to mean---and that cross-speaker comparison, speaker diarization, and semantic validation are essential safeguards against this category error. Computational social scientists should treat keyword-based modality scores as measurements of lexical co-occurrence, not rhetorical stance, and validate them accordingly.

Future work includes: (a) developing hybrid methods that preserve keyword interpretability while incorporating LLM-level semantic awareness, (b) applying the error typology to audit existing lexicon-based findings in the CSS literature, and (c) extending cross-model validation to additional model families to establish multi-LLM agreement as a practical reliability benchmark for domains where expert annotation is prohibitively expensive. The central lesson, however, is already clear: when a computational instrument produces a striking finding, the instrument itself deserves as much scrutiny as the speaker it claims to measure.

\subsection*{Data and Code Availability}
All source code, processed data, and interactive visualizations will be made public upon formal acceptance.


\end{document}